\documentclass[twoside,11pt]{article}

\usepackage{jair}
\usepackage{general}
\usepackage{enumerate}
\usepackage{graphicx}
\usepackage{amsfonts}
\usepackage{amsmath}
\usepackage{amssymb}
\usepackage{pstricks}
\usepackage{psfrag}
\usepackage{multirow}
\usepackage{subfigure}
\usepackage{url}

\newcommand{\marginlabel}[1]{\relax}

\begin{document}

\title{Numeric Input Relations for Relational Learning with Applications to
Community Structure Analysis}

\author{\name Jiuchuan Jiang \email jiangjiuchuan@163.com \\
\AND
\name Manfred Jaeger \email jaeger@cs.aau.dk \\
       \addr Department for Computer Science\\
       Aalborg University
     }


\maketitle


\begin{abstract}
Most work in the area of statistical relational learning (SRL) is focussed on discrete 
data, even though a few approaches for hybrid SRL models have been proposed that combine
numerical and discrete variables. In this paper we distinguish numerical random variables
for which a probability distribution is defined by the model from  numerical input 
variables that are only used for conditioning the distribution of discrete response variables. 
We show how numerical input relations can
very easily be used in the Relational Bayesian Network framework, and that existing
inference and learning methods need only minor adjustments to be applied in this
generalized setting. The resulting framework provides  natural relational extensions of 
classical probabilistic models for categorical data. We demonstrate the usefulness
of RBN models with numeric input relations by several examples.

In particular, we use the augmented RBN framework to define probabilistic models for 
multi-relational (social) networks in which the probability of a link between
two nodes depends on  numeric latent feature vectors associated with the 
nodes. A generic  learning procedure can be used to obtain
a maximum-likelihood fit of model parameters and latent feature values for
a variety of models that can be expressed in the high-level RBN representation.
Specifically, we propose a model that allows us to interpret learned 
latent feature values as community centrality degrees by which we can 
identify nodes that are central for one community, that are hubs between
communities, or that are isolated nodes. In a multi-relational setting, the model
also provides a characterization of how different relations are associated 
with each community.  


\end{abstract}

\section{Introduction}
\label{sec:introduction}

Statistical-relational learning (SRL) models   
have mostly been developed for discrete data (see~\cite{GetTas07,DeRFraKerMug08} for general overviews).
An important reason for this lies in the fact that inference for hybrid models combining discrete and continuous 
variables quickly lead to inference problems that consist of integration problems for which no closed-form 
solutions are available. 
Among the relatively few proposals for SRL frameworks with continuous variables are 
hybrid Markov Logic Networks~\cite{WanDom08}, hybrid ProbLog~\cite{GutJaeDeR10}, and hybrid 
dependency networks~\cite{RavRamDav15}. In these works
the complexity of the inference problem is addressed by focussing on approximate, sampling based 
methods~\cite{WanDom08,RavRamDav15}, or by imposing significant restrictions on the models, so that 
the required integration tasks for exact inference become solvable~\cite{GutJaeDeR10}. 


In the first part of this paper we first take a closer look at the semantic and statistical roles that
continuous variables can play in a probabilistic relational model. We arrive at a main distinction between numeric 
input relations, and numeric probabilistic relations, and we argue that for many modeling and learning problems involving 
numeric data,  only numeric input relations are needed. We then proceed to show how numeric input relations can 
be integrated into the Relational Bayesian Network (RBN) language~\cite{Jaeger97UAI}, with little or no cost in 
terms of algorithmic developments or computational complexity.  

The second part of the paper demonstrates by several examples and applications the usefulness of  
modeling with numeric input relations, and
the feasibility of the associated learning problems. First, a synthetic
environmental modeling example shows how RBNs with numeric input relations support natural and interpretable 
models that provide a relational extension for traditional statistical models (Section~\ref{sec:water}).

We then turn to community detection in (social) networks as our main application. 
Utilizing a general SRL modeling language allows us to encode a variety of probabilistic network models on a 
single platform with a single generic inference and learning engine. We use RBNs with numeric input relations
to encode probabilistic  models  with continuous latent features representing community structure. The SRL
framework makes it easy to develop models for multi-relational networks (a.k.a multiplex or multi-layer networks), 
where nodes are connected by more than one type of link.
In such networks, it will usually no longer be possible to reduce community structure detection to a form 
of graph partitioning~\cite{GirNew02}, because different relations may define a multitude of different, overlapping,
and partly conflicting community structures. We therefore propose a latent feature model that allows us to 
identify a number of communities with no restrictions on how communities are related in terms of
inclusion or disjointness. Furthermore, for each community we obtain a characterization of how they
are defined in terms of the given network relations, and we are able to define a probabilistic significance measure 
that ranks the detected communities in terms of their explanatory value.  Last but not least, we obtain for 
each node in the network, and each community, a  \emph{community centrality degree (ccd)}. Unlike most previously 
defined soft or fuzzy community membership degrees, these \emph{ccd} values are not normalized to sum up to
one over the different communities. They thereby allow us, for example,  to identify influential \emph{hub}
nodes~\cite{XuYurFenSch07} between communities (nodes with high centrality degree for multiple communities).

\section{SRL Models and Numeric Relations}
\label{sec:basics}

A SRL model defines a probability distribution over relational structures. A model can be 
instantiated over different input domains, which may consist only of a set of objects, or, 
more generally, a set of objects together with a set of known input relations. Thus, a SRL model
defines conditional probability distributions
\begin{equation}
  \label{eq:prmodel}
  P(\emph{IR}_{\emph{prob}}\mid \emph{IR}_{\emph{in}}, D, \boldtheta)
\end{equation}
where $D$\ ranges over a class of domains (usually the class of all finite sets), $\emph{IR}_{\emph{in}}$\ ranges over
interpretations over $D$\ of a set of input relations $R_{\emph{in}}$, and $\emph{IR}_{\emph{prob}}$\ ranges 
over interpretations of a set of probabilistic (or output) relations $R_{\emph{prob}}$. Interpretations
$\emph{IR}_{\emph{in}}$\ and $\emph{IR}_{\emph{prob}}$\ are given as value assignment to ground atoms $r(\boldd)$\ 
($\boldd\in D^{\emph{arity}(r)}$). In the discrete case, each relation $r$\ has an associated finite 
range of possible values. 
The distinction between input and probabilistic relation need not be explicitly defined in a given model.
Input relations can also be seen as probabilistic relations that are fully instantiated as evidence 
in a given inference or learning problem. Markov Logic Networks (MLNs)~\cite{RicDom06} are one prominent framework in 
which there is only such an implicit distinction between input and probabilistic relations.

\subsection{Hybrid SRL Models}

Hybrid SRL models allow the introduction of numeric relations, so that atoms $r(\boldd)$\ become
real-valued variables. Based on \eqref{eq:prmodel}, one can  distinguish 
numerical input and numerical probabilistic relations. 
To obtain a clearer view of the implications of this distinction, consider
a purely continuous, classical linear regression model:
\begin{equation}
\label{eq:linreg}
  P(Y\mid \boldX, \alpha, \boldbeta, \sigma) \approx \alpha+\boldX\cdot\boldbeta + N(0,\sigma^2)
\end{equation}
This model contains three different types of numerical variates: $Y$, the 
\emph{response variable}, is a random variable with a Gaussian distribution. 
$\boldX$, the \emph{predictor variables} may be random variables themselves, 
or they may be non-probabilistic inputs 
whose values have to be instantiated before inferences about $Y$\ can be made.
Finally, $\alpha,\boldbeta,\sigma$\ are \emph{parameters} of the model. 
The functional specification \eqref{eq:linreg} is completely symmetric for the
predictor variables $\boldX$\ and the parameters $\boldbeta$. The conceptual difference
between the two only becomes apparent when one considers repeated random samples 
$Y_1,\ldots,Y_n$. These samples would usually be drawn with varying values $X_1,\ldots,X_n$\ for 
the predictor variables, whereas the parameters $\boldbeta$\ remain constant.  

In SRL, data does not usually consist of iid samples, and one learns 
from a single observed pair $\emph{IR}_{\emph{in}},\emph{IR}_{\emph{prob}}$.
The distinction we can make in \eqref{eq:linreg} between $\boldX$\ and $\boldbeta$, therefore, 
is no longer supported. 
That means that in \eqref{eq:prmodel}  numeric input atoms $r(\boldd)$\ can be equally interpreted 
as predictor variables, or as object-specific parameters. 
Neither of these views requires to define $r(\boldd)$\ as a random variable with an associated 
probability distribution: as long as \eqref{eq:prmodel} is used purely as a conditional model, no prior
distribution for numeric input relations is needed.
Such a model will not define (posterior) probability distributions for numerical atoms, but still support maximum 
likelihood inference for the numerical atoms, which depending
on the interpretation of the input relations can be seen as MPE inference for unobserved 
predictor variables, or as estimation of object-specific parameters. 

The clear focus of Hybrid ProbLog~\cite{GutJaeDeR10} is to introduce numeric probabilistic 
relations. The language provides constructs to explicitly define distributions of numeric atoms as 
Gaussian with specified mean and standard deviation, for example. 

The nature of Hybrid MLNs~\cite{WanDom08} is a little less clear-cut, due
to the only implicit distinction between input and output relations
\footnote{For the following discussion we assume that the reader is familiar with  MLNs}. 
Hybrid MLNs extend standard MLNs by \emph{numeric properties} (which we 
can identify with numeric relations in our terminology) from which weighted numeric features can be 
constructed. Examples of weighted features that can be included in a hybrid MLN then are
\begin{eqnarray}
  \emph{distance}(X,Y) & \hspace*{10mm} w_1 \label{eq:mln1}\\
   - (\emph{length}(Z)-1.5)^2 & \hspace*{10mm} w_2 \label{eq:mln2}
\end{eqnarray}
A ground instance of a weighted feature contributes a weight to a possible world $x$\ (i.e. an interpretation
of all discrete and numerical relations over a given domain) that is equal to the value of the 
ground feature multiplied with the weight of the feature. The probability of $x$\ then is given by
$1/Z e^{W(x)}$, where $W$\ is the sum of weights from all groundings of all features, and $Z$\ a normalization
constant~\cite{WanDom08}. This definition, however, requires that a finite normalization constant $Z$\ can be found,
which means that $\int e^{W(x)}$\ must be finite, where the integral represents integration over all 
numeric properties, and summation over all discrete relations. This normalization is not possible, for example, for 
an MLN only consisting of the weighted feature \eqref{eq:mln1} with $w_1=1.0$. No probability distribution for the 
\emph{distance} property then is defined. For an MLN consisting
of \eqref{eq:mln2}, on the other hand, normalization is possible, and a Gaussian distribution for the
ground \emph{length} atoms is defined. When a hybrid MLN contains numeric properties that prevent normalization,
then no probabilistic inference for these properties is possible, and they either have to be instantiated to
perform probabilistic inference for other properties and relations, or one has to use the model for 
MPE inference tasks. In summary, hybrid MLNs support numeric probabilistic relations under the condition
that a finite normalization constant can be computed; otherwise they support numeric input relations for 
which no distribution is defined.  



\section{Numerical Input Relations in RBNs}
\label{sec:numrbn}

\subsection{Modeling}

The RBN language is based on \emph{probability formulas} that define
the probability $P(r(\bolda)=\emph{true})$\ for ground relational atoms 
$r(\bolda)$. The language of probability formulas is defined by a parsimonious
grammar that is based on the two main constructs of \emph{convex combinations}
and \emph{combination functions}. The following are two examples of the
convex combination construct. To improve the readability and understandability
of the formulas, we here use a modification of the original very compact syntax of~\cite{Jaeger97UAI}, 
and write convex combinations in the form of ``{\tt wif-then-else}''  statements
(``wif'' stands for ``weighted-if'').


\begin{eqnarray}
  \label{eq:probform1}
 & & 
\emph{P(heads(T) = true} )\ \leftarrow {\tt wif}\ \emph{fair}(T) \
     {\tt then }\  0.5 \ 
     {\tt else }\ 0.7  \\
\label{eq:probform2}
 & & 
\emph{P(cancer(X) = true} )\ \leftarrow\
    {\tt wif}\ 0.3  \ 
    {\tt then }\  \emph{genetic\_predisposition}(X)\ {\tt else }\  0.1
\end{eqnarray}

Formula \eqref{eq:probform1} defines the probability of a coin-toss to come up 
heads as 0.5 if a fair coin is tossed, and 0.7 otherwise. Here the formula in 
the {\tt wif}-clause is an ordinary Boolean condition. In formula \eqref{eq:probform2}
the  {\tt wif}-clause is a numerical mixture coefficient. The formula thereby defines
the probability that $X$\ has cancer as a mixture of a contribution coming from
a genetic predisposition (mixture weight 0.3), and a base rate of 0.1 (mixture 
weight 1-0.3). 

Generally a formula { {\tt wif} $A$\ \ {\tt then} $B$\ \ {\tt else} $C$}\ \ is evaluated over 
a concrete input domain to a probability value 
$\emph{val}({\tt wif}\ $A$\ \ {\tt then}\ $B$\ \ {\tt else}\ $C$)$\ as 
{ $  \emph{val}(A)\cdot\emph{val}(B) + (1-\emph{val}(A))\cdot \emph{val}(C)$}. 
There are two features in this modeling approach that make an integration of 
numerical input relations extremely easy: first, logical input relations already are
interpreted numerically: for example, $\emph{val}( \emph{fair}(T))$\ is defined 
as 0 or 1, depending
on whether $\emph{fair}(T)$\ is false or true. Second, according to the grammar of probability 
formulas, numerical constants and logical atoms are just different base cases for  
probability (sub-) formulas, which can be used interchangeably in the construction of 
more complex formulas. 

The generalization from Boolean to numerical relations, thus, is almost trivial: one
can just allow relational atoms $r(\bolda)$\ to evaluate to real values 
$ \emph{val}(r(\bolda))$\ in any range $[\emph{min},\emph{max}]$, where
$-\infty\leq \emph{min}\leq\emph{max}\leq \infty$\ depend on the intended
meaning of $r$.   
The only additional modification one has 
to make is to ensure that in the end probability formulas defining the probability for 
a Boolean response variable return values in the interval $[0,1]$. We do this by 
using the RBN combination function construct, which generally take multisets of 
(probability) values as inputs, and return a single probability value. In particular, we
can define \emph{logistic regression} as a RBN combination function. An example of 
a RBN model with numeric input relations and logistic regression combination function
then is:




\vspace{-4mm}

\begin{equation}
\label{eq:probform3}
\parbox{0.7\textwidth}{
\begin{tabbing}
\emph{P(cancer(A)=true)} $\leftarrow$ 
\=  {\tt  COMBINE}  \emph{intensity(R)}\\
\>    {\tt  WITH} \emph{l-reg} \\
\>     {\tt  FORALL} $R$\\
\>     {\tt  WHERE} $\emph{exposed}\,(A,R)$
\end{tabbing}
}
\end{equation}
\vspace{-4mm}

Again we here use a slightly more verbose version of the combination function syntax than
the original one. 
Formula  \eqref{eq:probform3} defines for a person $A$\ the probability of getting cancer using the logistic 
regression function applied to the set of all intensity values of radiation sources $R$\ that
$A$\ was exposed to. Thus, assuming that the numerical attribute \emph{intensity}, and the 
logical relation \emph{exposed} are known, the probability \emph{P(cancer(A))} evaluates to 
$exp(S)/(1+exp(S))$, where $S=\sum_{R:\emph{exposed}(A,R)}\emph{intensity}(R)$.

\subsection{Inference and Learning}
\label{sec:inflearn}

Probabilistic inference for RBNs with numerical input relations is no different from inference
in a purely Boolean setting. All inference approaches that have previously been used for 
RBNs (i.e., compilation to Bayesian networks or arithmetic circuits, importance sampling) can still be 
used without modifications.

\psfrag{prod}[c]{$\prod$}
\begin{figure}
  \centering
  \includegraphics[scale=0.35]{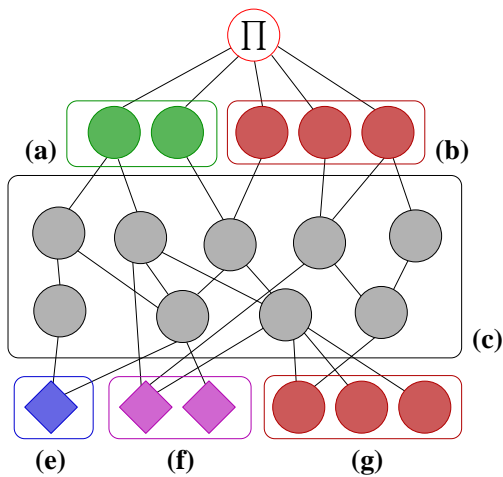}
  \caption{The likelihood graph. See text for key to (a)-(g)}
  \label{fig:likgraph}
\end{figure}
 
For learning the values of numerical relations, we use a slightly generalized version of the 
gradient graph that was introduced in~\cite{JaegerICML07}  for parameter learning in RBNs. The resulting
\emph{likelihood graph} data structure is illustrated in Figure~\ref{fig:likgraph}. 
The likelihood graph is a computational structure related to arithmetic circuits. 
Each node of the graph represents a function of the inputs in the bottom layer of the graph:
model parameters (e), values of ground atoms in the numerical input relations (f), and
truth values of ground probabilistic atoms that are unobserved in the data (g). 

The topmost layer of nodes in the graph corresponds to ground probabilistic atoms that are 
instantiated in the data (a), or that are unobserved and need to be marginalized out for the
computation of the likelihood (b) (there is a one-to-one correspondence between the nodes in 
(b) and (g)). The function associated with the ground atom nodes of this layer is the probability of the
atom, given the current parameter settings, and instantiations of the ground atoms in (g).  

The nodes in the intermediate layers (c) represent sub-formulas of the probability formulas for 
the ground atoms in (a) and (b). Finally, the root node represents the product over all 
nodes in (a) and (b), and thus represents the likelihood of the joint configuration of probabilistic
atoms consisting of the observed values for (a), and the current setting at the nodes (g) for the 
atoms in (b). 

The likelihood graph is constructed in  a top-down manner by a recursive decomposition of 
the probability formulas. 
In this decomposition also sub-formulas will be encountered that 
have a constant value, and do not depend on any parameters or unobserved atoms. These sub-formulas 
are not represented explicitly by nodes in the graph and are not 
decomposed further. Their constant value is directly assimilated into the 
function computation at their parent nodes.

The likelihood graph supports  computation of the likelihood values and the
gradient of the likelihood function with respect to the numerical parameters (e) and (f). 
These computation are linear in the number of edges of the graph. Based on these 
elementary computations, the likelihood graph can be used for parameter learning via 
gradient ascend, marginalization over unobserved atoms via MCMC sampling (mainly used 
when learning from incomplete data), and MAP inference for unobserved probabilistic atoms.
For parameter learning, we perform multiple restarts of gradient ascend with random
initializations for the nodes (e), (f), (g).

\section{Examples: Standard Logistic Regression}

We now demonstrate the usefulness of RBN models with numerical input relations for 
practical modeling and learning problems, and the feasibility of learning via 
likelihood graph based gradient ascent. 
In this section we present examples that demonstrate the use of the logistic regression
model in our relational framework for constructing models that closely follow 
conventional and interpretable statistical modeling approaches.

\subsection{Propositional: Cancer Data}
\label{sec:classic}

In a first experiment we test whether standard ``propositional'' logistic regression is 
properly embedded in our relational framework. For this we use a very small dataset containing 
data on 27 cancer patients that was originally
introduced in~\cite{Lee1974}, and which is often used as a standard example for logistic regression.
We use a simplified version of the dataset given in~\cite[Table 5.10]{Agresti02}, which contains
a single numerical predictor variable \emph{LI}, and a binary response variable indicating 
whether the cancer is in \emph{remission} after treatment. A standard logistic regression model 
for predicting \emph{remission} is represented by the probability formula

\begin{equation}
\label{eq:remission}
\emph{P(remission(A)=true)} \leftarrow 
{\tt  COMBINE}\ \alpha + \beta\cdot \emph{LI(A)}\ \ {\tt  WITH}\ \emph{l-reg}. 
\end{equation}

The combination function construct in this formula is somewhat degenerate, since it here effects no
combination over a multiset of values, and simply reduces to the application of the logistic 
regression function to the single number $\alpha + \beta\cdot \emph{LI(A)}$.

\begin{figure}[t]
  \centering
\includegraphics[scale=0.4]{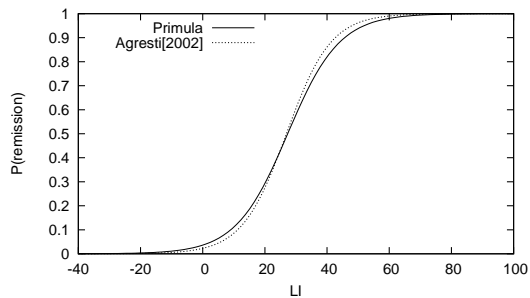}
  \caption{Model learned for \emph{remission} probability}
\label{fig:remission}
\end{figure}

Figure~\ref{fig:remission} shows the probability of the response variable as a function of the 
predictor variable for the parameters $\alpha,\beta$\ learned from the RBN encoding \eqref{eq:remission}, and
for the parameters given in~\cite{Agresti02} (which were fitted using the SAS statistics toolbox). 
Clearly, our gradient ascent approach using the likelihood graph here yields results that are 
compatible with standard approaches to logistic regression.

\subsection{Relational: Water Network}
\label{sec:water}

In this section we consider a toy model for the propagation of pollution in a river 
network. This example demonstrates the ability to integrate into our relational 
modeling framework standard logistic 
response models based on meaningful and interpretable predictor relations.

Input domains for this model consist of measuring stations in a river network that
measure whether the river is \emph{polluted} or not. Stations are related by the Boolean
\emph{upstream} relation, denoting that one station is directly upstream of another (i.e., 
without any other stations in-between). For any pair of stations in the \emph{upstream} relation,
there also is a numerical relation \emph{invdistance} containing the inverse of the distance
between the two stations.

\begin{figure}
\psfrag{S1}[c]{\scalebox{0.6}{$S_1$}}
\psfrag{S2}[c]{\scalebox{0.6}{$S_2$}}
\psfrag{S3}[c]{\scalebox{0.6}{$S_3$}}
\psfrag{S4}[c]{\scalebox{0.6}{$S_4$}}
\psfrag{S5}[c]{\scalebox{0.6}{$S_5$}}
\psfrag{S6}[c]{\scalebox{0.6}{$S_6$}}
\psfrag{S7}[c]{\scalebox{0.6}{$S_7$}}
\psfrag{S8}[c]{\scalebox{0.6}{$S_8$}}
\psfrag{S9}[c]{\scalebox{0.6}{$S_9$}}
\psfrag{S10}[c]{\scalebox{0.6}{$S_{10}$}}
\psfrag{S11}[c]{\scalebox{0.6}{$S_{11}$}}
\psfrag{S12}[c]{\scalebox{0.6}{$S_{12}$}}

  \centering
  \includegraphics[scale=0.2]{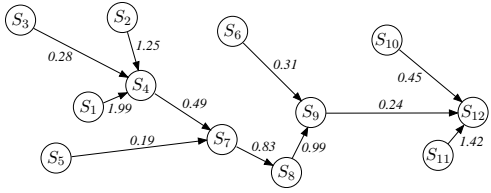}
  \caption{Water Network Input Domain}
  \label{fig:waternetw}
\end{figure}

The outermost wif-then-else construct in lines 1,2,10 of the model of Table~\ref{tab:river} 
defines the \emph{polluted} attributed as a mixture of two 
factors: first, there is a base probability of $(1-0.6)\cdot 0.2=0.08$\ for pollution to 
occur regardless of pollution already being measured upstream. Second, lines 2.-9. contain a 
propagation model of pollution that is measured at one or several upstream stations. 
This  probability sub-formula computes the expression
 
\begin{displaymath}
  \emph{l-reg}\, (\alpha +\beta \hspace{-8mm}
\sum_{V: \begin{array}[t]{l} \scriptstyle \emph{upstream}(V,S) \& \\ [-1mm]
                               \scriptstyle \emph{polluted}(V) 
            \end{array}}  
\hspace{-8mm}
\emph{invdistance}(V,S)\, ).
\end{displaymath}

\begin{table}
\caption{Pollution model \label{tab:river}}
\begin{tabbing}
1.\hspace{5mm} \emph{polluted}\, $(S)\leftarrow$  \= {\tt WIF }\hspace{2mm}   0.6 \\
2. \>  {\tt THEN}  \= {\tt COMBINE} \= $\alpha$, \\
3. \> \>  \>           {\tt COMBINE} \= {\tt  WIF}  \emph{polluted(V)}\\
4.  \> \> \> \>                    {\tt THEN} $\beta$ * \emph{invdistance(V,S)}\\
5.  \> \> \> \>                    {\tt ELSE}  0.0\\
6.  \> \> \>            {\tt  WITH} sum\\
7.  \> \> \>             {\tt FORALL} \emph{V} \\
8.  \> \> \>           {\tt  WHERE} \emph{upstream(V,S)}\\
9.  \>  \>     {\tt WITH} l-reg\\
10.  \> {\tt ELSE } 0.2;\\
\end{tabbing}
\end{table}

\begin{figure}
  \centering  
   \includegraphics[scale=0.4]{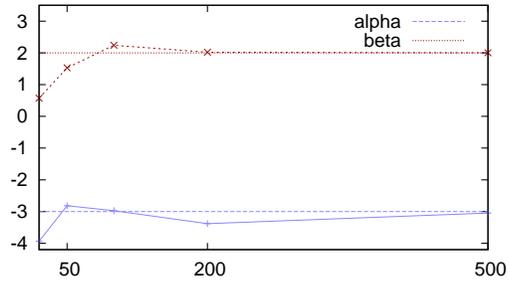}

(a)

  \includegraphics[scale=0.4]{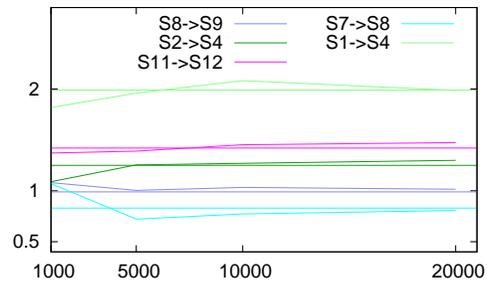}

(b)

 \includegraphics[scale=0.4]{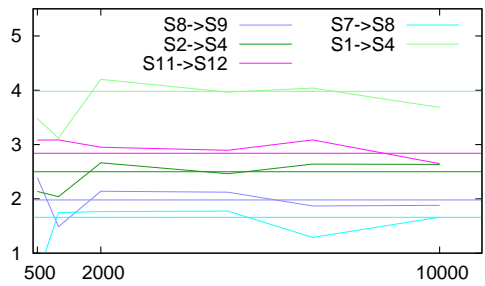}

(c)
  \caption{Convergence of estimates in water network example}
  \label{fig:waterparams}
\end{figure}

Generalizing from the baseline example of Section~\ref{sec:classic} we investigate
whether the parameters of the model can still be identified from independent samples of
the \emph{polluted} attribute. To this end we sample $N$\ independent joint instantiations 
of the \emph{polluted} attribute for the 12 measuring stations of the domain in 
Figure~\ref{fig:waternetw} with parameters $\alpha=-3$\ and $\beta =2$, and the values of the
\emph{invdistance} relation as shown in Figure~\ref{fig:waternetw}.  All experiments are performed using
20 random restarts. In the first experiment the values of the \emph{invdistance} relation
are fixed at their true values of Figure~\ref{fig:waternetw}, and we only learn the values 
of $\alpha,\beta$. Figure~\ref{fig:waterparams} (a) shows the learned values for increasing sample sizes 
$N=20,50,200,500$. 
Clearly, quite accurate estimates for $\alpha,\beta$\ are already obtained from relatively
small sample sizes. 

In a second experiment, $\alpha,\beta$\ are fixed at their true values, and the
values of the \emph{invdistance} relation are learned.  Figure~\ref{fig:waterparams} (b) shows
the convergence of the estimates for the \emph{invdistance} values of five  different
pairs of neighboring measuring stations. Again, the true values are consistently learned. The
required sample size is much larger than for $\alpha,\beta$, because a single sample 
only contains relevant information for the estimation of $\emph{invdistance}(S_i,S_j)$\ when
$\emph{polluted}(S_i)$\ is true in that sample.

In a third experiment, both $\alpha,\beta$\ and the \emph{invdistance} relations are learned.
In this setup no convergence to the true values can be expected, since the parameters are not
jointly identifiable: for any given setting of the parameters $\alpha,\beta,\emph{invdistance}()$, 
one obtains  equivalent solutions in the form $\alpha,\lambda\cdot\beta,\emph{invdistance}/\lambda$.
For this reason, we compare the products $\beta\cdot\emph{invdistance}()$\ for both the parameters
in the generating and learned model. Figure~\ref{fig:waterparams} (c) shows these products for the
same pairs of stations as in (b). The convergence here shows that even if the exact values of 
the parameters cannot be learned, a probabilistically equivalent model is learned (the learned 
value of the parameter $\alpha$\ also converges to the true value -3.0). 

 \begin{table}
   \centering
   \caption{Size of likelihood graph and learning times}
   \label{tab:complwater}
   \begin{tabular}{c|rrrrr}
& \multicolumn{5}{c}{$N=$} \\
   & \multicolumn{1}{c}{1000} & \multicolumn{1}{c}{2000} & 
     \multicolumn{1}{c}{4000} & \multicolumn{1}{c}{8000} & \\ \hline
   \multirow{2}{*}{\# nodes} & 7095 & 14611 & 29331 & 59325 & \\
   & 28995 & 79500  & 115874& 318539 & \\ \hline
   \multirow{2}{*}{construction (s)} & 0.37 &  0.63 & 1.17 & 2.69 & \\
   & 0.87 & 1.25  & 2.96 & 5.86 & \\ \hline
   \multirow{2}{*}{time/restart (s)} & 0.94 &  3.38 & 6.61 & 12.45 & \\
   & 13.26 & 36.92  & 59.64 & 117.30 & 
  \end{tabular}
 \end{table}

Table~\ref{tab:complwater} shows the size of the likelihood graph, the time for construction, 
and the average time per restart for the gradient ascent optimization. For different values
of the sample size $N$, these numbers are given for the case where we only learn the 
relation \emph{invdistance}() (top entry in each cell of the table), and the case where 
we learn $\alpha,\beta$\ and \emph{invdistance}() (bottom entry). The likelihood graph is 
significantly larger when also learning $\alpha,\beta$, because here more sub-formulas of the
instantiated model depend on unknown parameters, and therefore can not be pruned in the 
construction.

\section{Application: Community Structure in Multi-Relational Networks}
\label{sec:commstruc}

\begin{figure}
  \centering
  \includegraphics[scale=0.5]{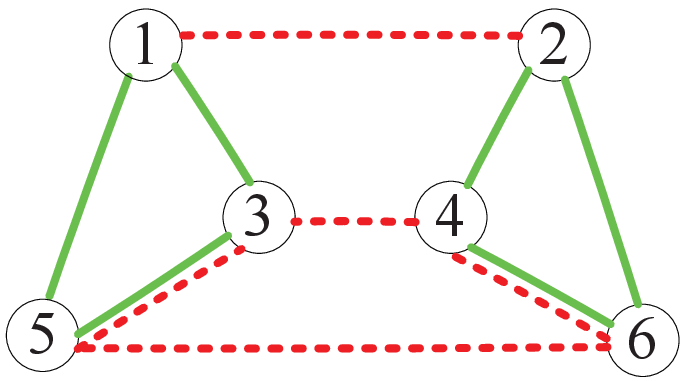}
  \caption{Multi-relational community structure}
  \label{fig:introex1}
\end{figure}

We will now apply learning of numeric input relations in RBNs for community
structure analysis in multi-relational networks.
Figure~\ref{fig:introex1} shows a small network with 6 individuals connected by two 
different types of (undirected) links. Considering only the green (solid) link relation,
one would identify \{1, 3, 5\} and \{2, 4, 6\} as communities, whereas the red (dashed) link 
relation points to communities \{1, 2\} and \{3, 4, 5, 6\}. Moreover,  the community
structure  \{1, 3, 5\}, \{2, 4, 6\},  would indicate that the red links are representing
an antagonistic relationship that is more likely to exist in between communities, than within communities.
Considering both links simultaneously, and assuming both are positive indicators of 
communities, one may also consider \{3,5\} and \{4,6\} as the most clearly defined
communities, to which 1 and 2 are more loosely connected. 

This tiny example illustrates how multiple relations can lead to a rather complex community
landscape, with multiple possible views and interpretations. Even though multi-relational
networks occur naturally, research on community detection has very much focused on the 
single-relational case. Proposals for dealing with multi-relational networks often 
consist of reductions to the single-relational setting, either by aggregating all relations
into a single weighted relations~\cite{CaietAl05,KivetAl14}, or by aggregating results from
community detection performed for each relation separately~\cite{BerPinCal13}.

In Section~\ref{sec:commmultrel} we will propose a latent feature model that takes all relations as input
in a non-aggregated form, and returns multiple communities along with a characterization of
how the different communities are correlated with the relations.

\begin{figure}
  \centering
  \includegraphics[scale=0.5]{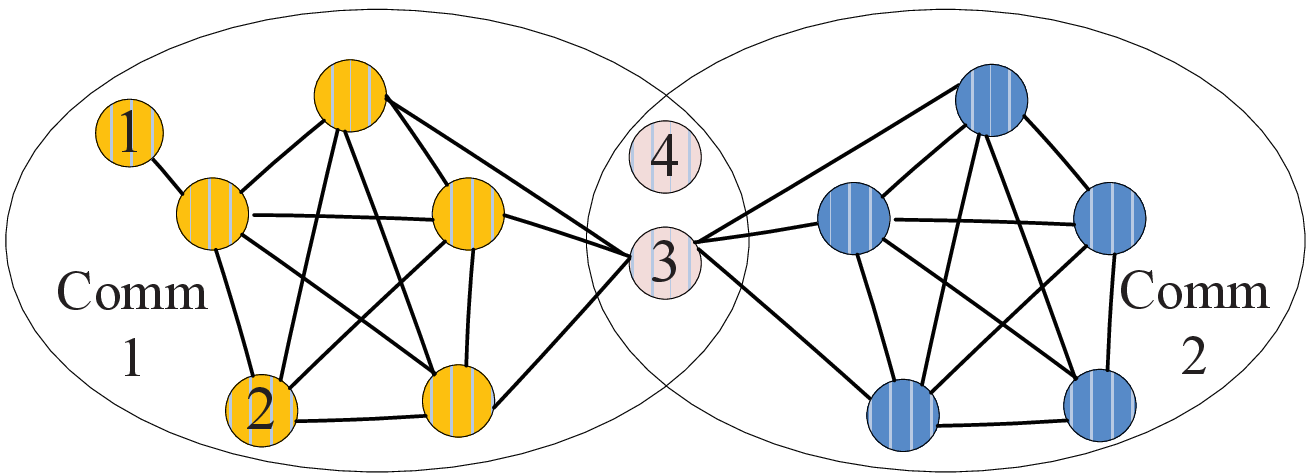}
  \caption{Multi-relational community structure}
  \label{fig:introex2}
\end{figure}

Figure~\ref{fig:introex2} shows a single-relational network with a relatively clear two-community
structure. The two nodes 3 and 4, however, are perfectly ambiguous with regard to their 
community membership. Most existing soft clustering methods would give both nodes equal 
membership degrees of 0.5 for both communities. However, clearly it is desirable to be able 
to distinguish node 3, which is well connected to both communities, and which for information
diffusion purposes would be the most influential node in the network~\cite{DomRic01}, from node 4,
which is completely isolated. Nodes 1 and 2 are both very strongly associated with the 
community on the left. However, instead of assigning a membership degree close to 1.0 to 
both of them, it will be more informative to assign a higher membership degree to node 2 than 
to node 1, so that the membership degree also reflects the centrality of the nodes for the
communities. In our model, the learned values of latent numeric relations can be interpreted as  
\emph{community centrality degrees}, that reflect the degree of connectivity of a node with 
all communities.

\subsection{Community Centrality Degrees}

In this section we first consider the single-relational case to explore models for 
learning community membership degrees that satisfy the desiderata outlined in the preceding 
section. We introduce a numerical binary relation $u(V,C)$, whose  arguments are a node
$V$, and  a community $C$. 
The relation $u$\ is constrained to be non-negative.
We can then define the following probabilistic model:
\begin{equation}
  P(\emph{link}(V,W)) = 
\left\{ 
  \begin{array}{ll}
    0 & V=W \\
    e^S/(1+e^S) & V\neq W 
  \end{array}
\right. 
\label{eq:singlerelmodel1}
\end{equation}
where
\begin{equation}
  S=\alpha+\sum_{C:\emph{community}(C)}u(V,C)\cdot u(W,C)
\label{eq:singlerelmodel2}
\end{equation}
and $\alpha$\ is a real-valued constant (the intercept, in the language of log-linear models). 
This model is  quite straightforward, and closely related to other models for link prediction
(e.g. for recommender systems) in which the affinity of objects to be connected by a link 
is measured by the inner product of latent feature vectors 
associated with the objects.  We note that in contrast to structurally similar probabilistic latent semantic
models~\cite{Hofmann04} the variables $u(V,C),u(W,C)$\ have no semantics as (conditional) 
probabilities, and \eqref{eq:singlerelmodel1},\eqref{eq:singlerelmodel2} is not a
mixture model with the communities as hidden mixture components. The model is readily encoded
as a RBN. The observed links in a  network with node set ${\cal N}$\ then define the likelihood function
\begin{multline}
  \label{eq:graphll}
  L(\alpha, u) =\hspace{-5mm} \prod_{V,W\in {\cal N}:\emph{link(V,W)=\emph{true}}}\hspace{-10mm} P( \emph{link}(V,W)) 
 \prod_{V,W\in {\cal N}:\emph{link(V,W)=\emph{false}}}\hspace{-10mm}(1-P( \emph{link}(V,W)) )
\end{multline}
The generic learning method described in Section~\ref{sec:inflearn} can be used to fit 
the model parameters $\alpha\cup \{u(V,C)\mid V,C: \emph{node}(V),\emph{cluster}(C)\}$. 

\psfrag{uc1}{$u(\cdot,C_1)$}
\psfrag{uc2}{\rotatebox{90}{$u(\cdot,C_2)$}}

\begin{figure*}[t]
 \centering
 \includegraphics[scale=.18]{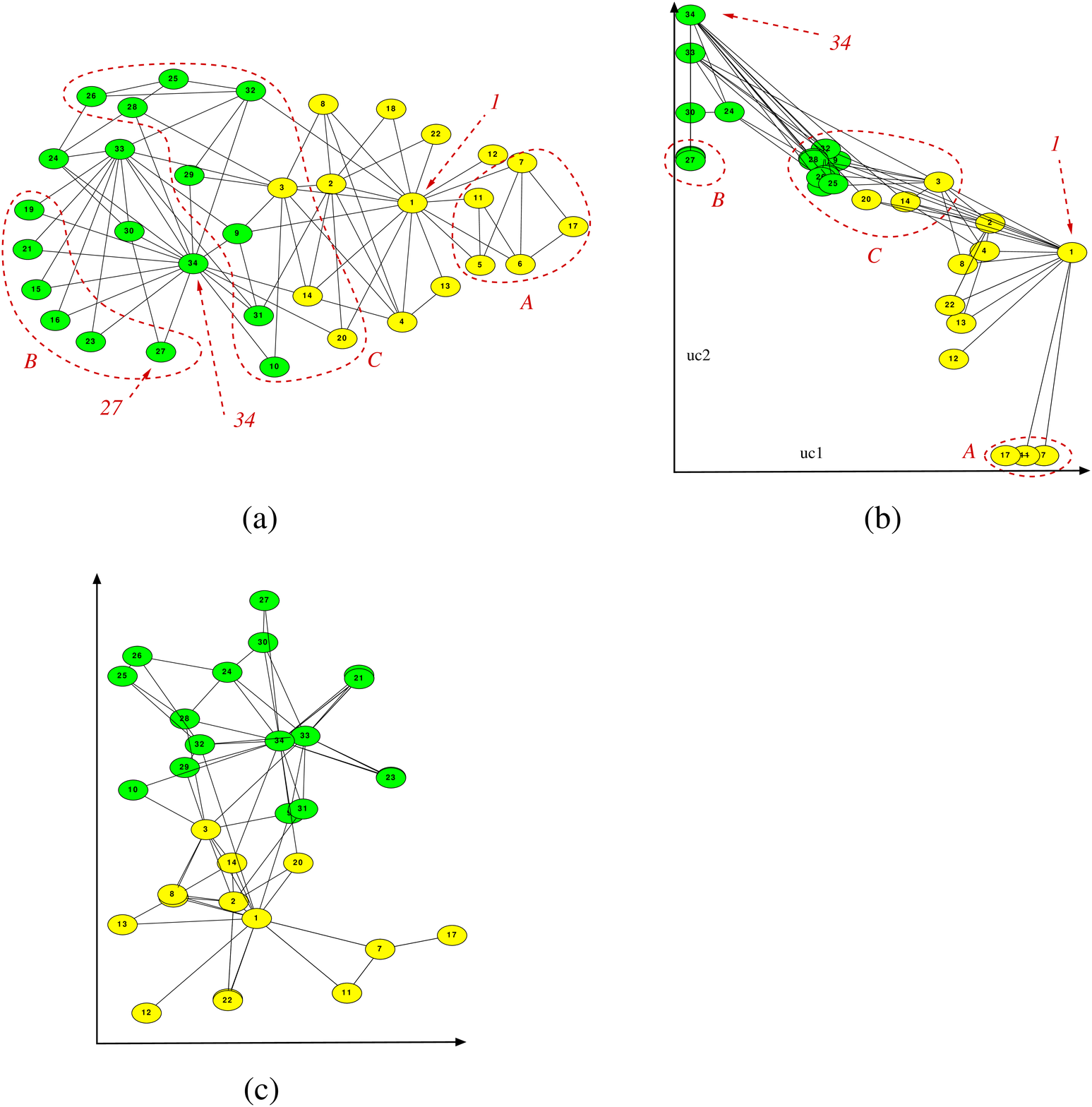}
 \caption{Zachary network}
   \label{fig:zachary}
 \end{figure*}

We  applied this model to the well known Zachary Karate Club network depicted in 
Figure~\ref{fig:zachary} (a), where the node colors represent the known ``ground truth''
communities in this network~\cite{Zachary77}. 
Figure~\ref{fig:zachary} (b) illustrates the learned $u$-values when the model 
is instantiated for 2 communities $C_1,C_2$.  Nodes $V$\ are plotted in 2-dimensional space
according to their $u(V,C_1),u(V,C_2)$\ values. Node colors still represent the ground truth.
Some individual nodes, and groups of nodes, are marked correspondingly in 
 Figure~\ref{fig:zachary} (a)  and (b).  
We first observe that the nodes 1 and 34 with maximal $u(\cdot,C_1)$\ and $u(\cdot,C_2)$-values 
are  central nodes of their respective communities. In contrast, the node groups $A$\ and $B$\ are 
well-connected with their own communities, but separated from the other community. $C$\ is a large
group of nodes with $u(\cdot,C_1)$\ and $u(\cdot,C_2)$-values of similar magnitude. All nodes in 
this group can be seen as potential hub nodes between the two communities, but node 3 with 
the highest sum $u(\cdot,C_1) + u(\cdot,C_2)$\ is most clearly identified as a well-connected
hub  between the communities. 

Two further observations are worthwhile making: the fact that some  $u$-values of zero have been learned
indicates that allowing negative $u$-values could lead to higher likelihood scores. However, for 
the purpose of interpretability of the results, imposing the non-negativity constraint for $u$\
still seems beneficial. Second, for nodes that are pairwise structurally indistinguishable (all nodes 
other than node 27 in group $B$) identical $u$-values were learned. This, obviously, is highly
desirable, and supports both the adequacy of the probabilistic model, and the effectiveness of 
the optimization procedure (which, starting from random initial values, could be feared to 
get stuck in local optima with non-identical values). 

We compare the results obtained with model \eqref{eq:singlerelmodel1},\eqref{eq:singlerelmodel2}
with a slight modification of the \emph{distance model} proposed in~\cite{HofRafHan02}. This model
is given by \eqref{eq:singlerelmodel1} in conjunction with 
\begin{equation}
  S=\alpha+\sum_{c:\emph{community}(C)}(u(V,C) -  u(W,C))^2.
\label{eq:singlerelmodel3}
\end{equation}
Thus, the log-odds of the \emph{link} probability now depend on the squared Euclidean distance
between the latent feature vectors. This model, too, is readily encoded by a RBN, and the learned
$u$-values are visualized in Figure~\ref{fig:zachary} (c). The positions of the nodes in the 
latent space here are not interpretable as community centrality degrees, and (in line with the 
motivation given by the authors for this model)  rather are suitable as node-coordinates for
graph visualization and plotting. The model  \eqref{eq:singlerelmodel1},\eqref{eq:singlerelmodel3}
also achieves a much lower log-likelihood of -245 than the model 
\eqref{eq:singlerelmodel1},\eqref{eq:singlerelmodel2}, which achieves a log-likelihood of -157. 
The baseline model that does not contain any latent feature vectors $u$, and only the $\alpha$\ parameter
is fitted (i.e., a fitted Erd\"os-R\'enyi model), achieves a log-likelihood score of -452.  

The likelihood graphs for the models \eqref{eq:singlerelmodel1},\eqref{eq:singlerelmodel2} and 
\eqref{eq:singlerelmodel1},\eqref{eq:singlerelmodel3} contain 5679  and 10235 nodes,
respectively. 
The construction times for the graphs are around 0.1s and 0.8s, respectively. The times per restart
of the learning procedure was around 31s for both models. The increased computation time per 
gradient computation in the larger graph for the second model was offset by a smaller number 
of iterations required until convergence. The obtained results are quite robust: solutions with
very similar likelihood scores and structures as the ones shown in Figure~\ref{fig:zachary} are 
usually obtained as the highest-scoring solutions within 3-5 restarts.

\subsection{Communities in Multi-Relational Networks}
\label{sec:commmultrel}

\begin{figure*}
  \centering
  \subfigure[]{\includegraphics[width=0.3\textwidth]{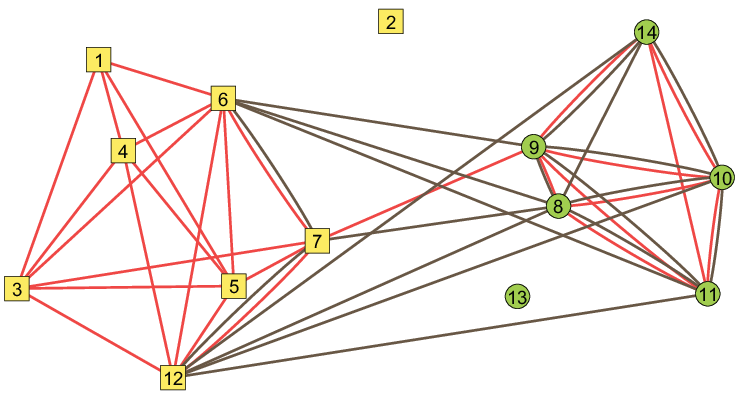}
\label{fig:wiringnetworkPositive}}
\hspace{0.5cm}
\subfigure{\includegraphics[width=0.15\textwidth]{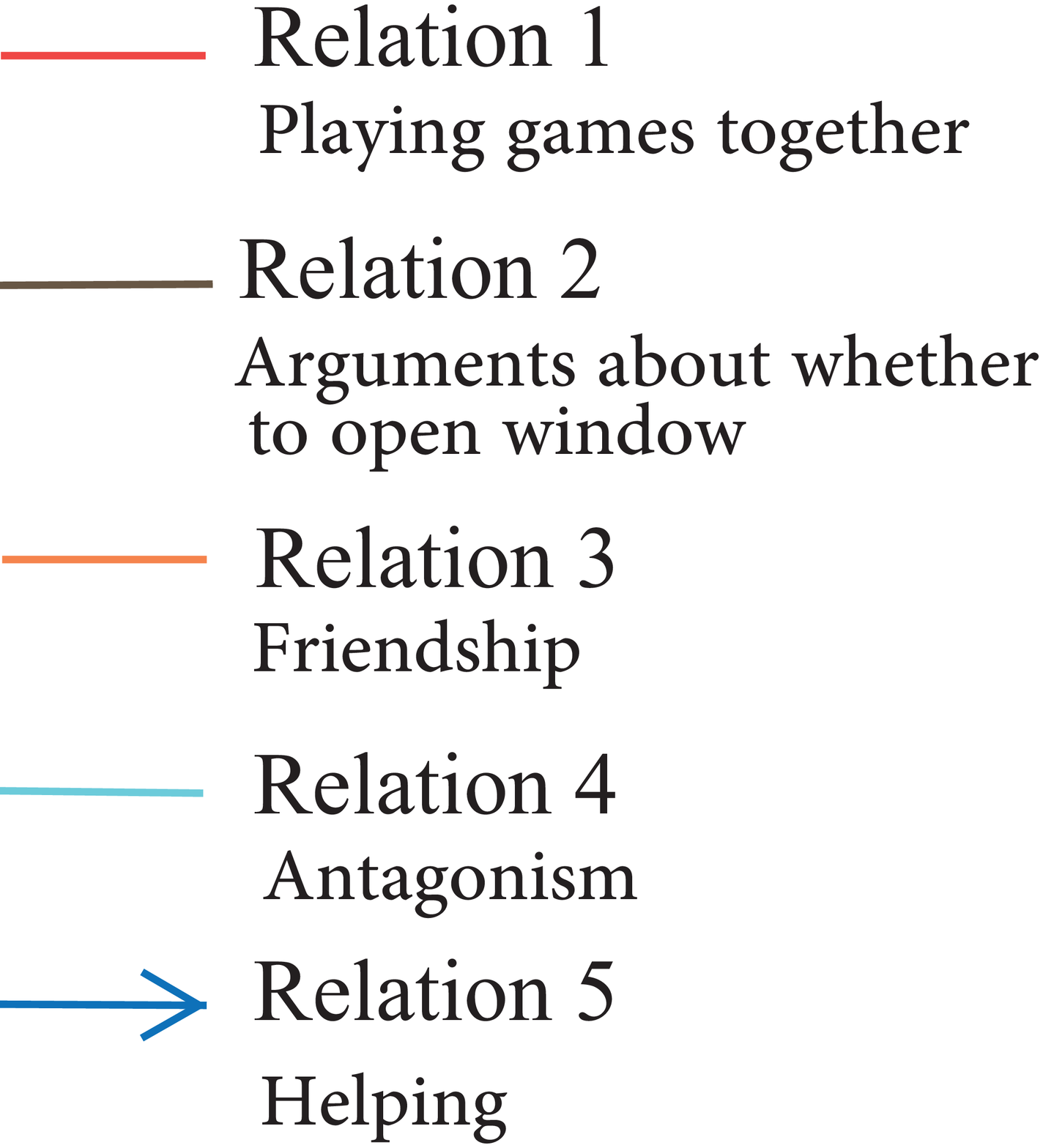}
\label{fig:key}}
\hspace{0.5cm}
\setcounter{subfigure}{1}  
\subfigure[]{\includegraphics[width=0.3\textwidth]{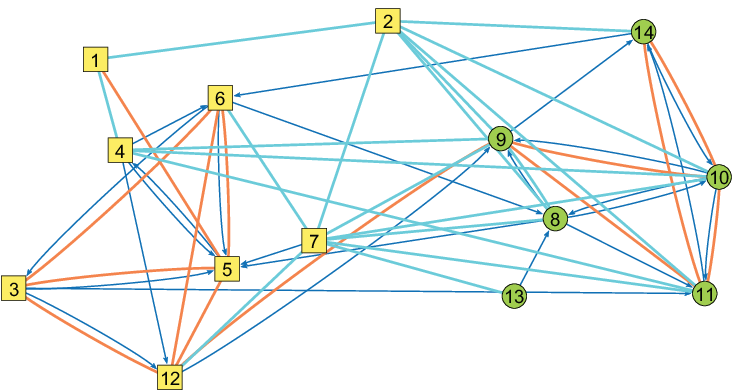}
\label{fig:wiringnetworkNegative}}
\caption{The Wiring Room multiplex network}
 \label{fig:wiring}
\end{figure*}

We now generalize the model \eqref{eq:singlerelmodel1},\eqref{eq:singlerelmodel2} to 
multi-relational networks. For a network containing $K$\ relations 
$\emph{link}_i$\ ($i=1,\ldots,K$), 
we introduce $K$\ new numeric attributes $t_i(C)$\ on the cluster objects of the domain. 
The values of the $t_i$\ are unconstrained. The intention is that $t_i(C)$\ measures whether 
the existence of links of type $i$\ is positively ($t_i>0$) or negatively ($t_i<0$) correlated with 
membership in cluster $C$. We now define the probability $P(\emph{link}_i(V,W))$\ using 
 \eqref{eq:singlerelmodel1} in conjunction with
 \begin{equation}
   \label{eq:multrbnmodel}
   S_i=\alpha_i+\sum_{c:\emph{community}(C)}u(V,C)\cdot u(W,C)\cdot t_i(C).
 \end{equation}
Given an observed network with $K$\ different \emph{link} relations, we have to fit 
the model parameters $\{\alpha_i\mid i=1,\ldots,K\}\cup
\{u(V,C)\mid V,C: \emph{node}(V),\emph{cluster}(C)\}\cup
\{ t_i(C)\mid C,i: \emph{cluster}(C), i=1,\ldots,K   \}$.
This model is clearly not identifiable: for a given parameterization, 
multiplying all $t_i$-values with a factor $c>0$,
and dividing all $u$-values by $\sqrt{c}$\ leads to an equivalent parameterization.
Absolute numeric values of the fitted parameters are  therefore not significant, but relative 
magnitudes of values can still identify community structure.

{
\psfrag{C1}{\raisebox{-1mm}{\Huge $C_1$}}
\psfrag{C2}{\raisebox{-1mm}{\Huge $C_2$}}
\psfrag{C3}{\raisebox{-1mm}{\Huge $C_3$}}
\psfrag{C4}{\raisebox{-1mm}{\Huge $C_4$}}
\psfrag{0}[c]{\raisebox{-2mm}{\scalebox{3.0}{0}}}
\psfrag{1}[c]{\raisebox{-2mm}{\scalebox{3.0}{1}}}
\psfrag{2}[c]{\raisebox{-2mm}{\scalebox{3.0}{2}}}
\psfrag{3}[c]{\raisebox{-2mm}{\scalebox{3.0}{3}}}
\psfrag{4}[c]{\raisebox{-2mm}{\scalebox{3.0}{4}}}
\psfrag{5}[c]{\raisebox{-2mm}{\scalebox{3.0}{5}}}
\psfrag{6}[c]{\raisebox{-2mm}{\scalebox{3.0}{6}}}
\psfrag{7}[c]{\raisebox{-2mm}{\scalebox{3.0}{7}}}
\psfrag{8}[c]{\raisebox{-2mm}{\scalebox{3.0}{8}}}
\psfrag{9}[c]{\raisebox{-2mm}{\scalebox{3.0}{9}}}
\psfrag{10}[c]{\raisebox{-2mm}{\scalebox{3.0}{10}}}
\psfrag{11}[c]{\raisebox{-2mm}{\scalebox{3.0}{11}}}
\psfrag{12}[c]{\raisebox{-2mm}{\scalebox{3.0}{12}}}
\psfrag{13}[c]{\raisebox{-2mm}{\scalebox{3.0}{13}}}
\psfrag{14}[c]{\raisebox{-2mm}{\scalebox{3.0}{14}}}

\begin{figure}[b]
  \centering
  \includegraphics[angle=-90,scale=0.2]{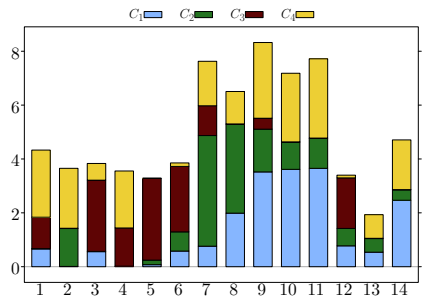}
  \caption{Learned $u$\ values}
  \label{fig:wiringplotsus}
\end{figure}

We apply the model to the multi-relational \emph{wiring room} network~\cite{BreBooAra75} depicted in 
Figure~\ref{fig:wiring}. The network consists of 14 nodes connected by 5 distinct relations.
For better visibility, the relations here are displayed in two groups. Out of the 5 relations,
3 represent positive relationships, 1 is antagonistic, and 1 (``arguments about opening a window'') 
potentially ambivalent. Relation 5 is directed, the others undirected. 
The coloring of the nodes represent a community structure found
in~\cite{BreBooAra75} for this network. 

\psfrag{C1}{\raisebox{-4mm}{\Huge $C_1$}}
\psfrag{C2}{\raisebox{-4mm}{\Huge $C_2$}}
\psfrag{C3}{\raisebox{-4mm}{\Huge $C_3$}}
\psfrag{C4}{\raisebox{-4mm}{\Huge $C_4$}}

\begin{figure}[b]
  \centering
  \includegraphics[angle=-90,scale=0.2]{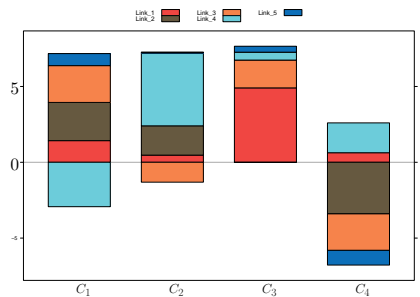}
  \caption{Learned $t_i$-values}
  \label{fig:wiringplotsthetas}
\end{figure}
}

Using 4 clusters, we learn  $u(\cdot,C_i)$-values for the 14 nodes as shown in
Figure~\ref{fig:wiringplotsus} , and $t_i(\cdot)$-values for the 4 communities as
shown in Figure~\ref{fig:wiringplotsthetas}.
The values $u(\cdot,C_1)$\ (light blue in Figure~\ref{fig:wiringplotsus}) identify 
nodes 9,10,11 as central nodes of community $C_1$, to which 8 and 14 also are strongly 
associated. This very much coincides with the original green community. According to the 
$t_i(C_1)$-values of  Figure~\ref{fig:wiringplotsthetas}, membership in this community 
is most positively associated with relations 2 and 3, and to a lesser extent 1 and 5.
Relation 4 is clearly negatively associated with this community. Similarly, there is a good
correspondence between community $C_3$, and the original yellow community. According to the
$t_i$-values, this community is most clearly associated with relations 1 and 3. 
Community $C_2$\ considers relation 4 as strongly positive, and thereby provides a 
non-standard view on the community structure of this network, with nodes 7,8 the centers
of this community. Finally, community  $C_4$\ also considers relations 4 as positive, but 
unlike for $C_2$, there is a negative association with relation 2. 

The likelihood graph here contained 7361 nodes. The reported result 
is the best obtained in 10 random restarts of the learning procedure, where one 
restart took about 1 minute to compute.

\subsection{Community Significance Measure}

It is highly desirable that a method not only returns the requested number of communities,
but also provides a measure of the significance, or validity, of each community. On the basis
of our probabilistic model, we obtain such a measure in terms of the explanatory value
that a community provides for the observed network structure, where explanatory value
is formalized by the likelihood gain obtained by including community information into the
model. 

Specifically, to measure the explanatory value of community $C_k$\ defined by the
$u(\cdot,C_k)$-values of all nodes, we consider the model given by \eqref{eq:singlerelmodel1} 
in conjunction 
 \begin{equation}
   \label{eq:likgainmodel}
   S_i=\alpha_i+u(V,C_k)\cdot u(W,C_k)\cdot t_i(C_k).
 \end{equation}
In this model, we now keep the previously learned $u(\cdot,C_k)$\ values fixed, and
re-learn the parameters $\{\alpha_i\mid i=1,\ldots,K\}\cup \{ t_i(C_k)\mid i=1,\ldots,K \}$.
As a baseline we take the Erd\"os-R\'enyi log-likelihood obtained when only fitting the
$\alpha_i$-values in a model without communities. We then define the \emph{likelihood gain}
obtained from community $C_k$\ as the log-likelihood obtained by \eqref{eq:likgainmodel}, minus
the   Erd\"os-R\'enyi baseline. 

For the communities identified for the Wiring Room network we obtain the following 
likelihood gain values: $C_3: 71.4$, $C_1 :62.0$, $C_2:39.5$, $C_4:14.4$. Thus, the ranking 
obtained by the likelihood gain scores reflects quite well the intuitive evaluation of 
the communities in terms of  interpretability.

\subsection{Incomplete Network Data}

An important benefit of using probabilistic models for network analysis is the ability to 
handle incomplete information: for  the likelihood function \eqref{eq:graphll} it is not
required that for every pair of nodes $V,W$\ the true/false status of the \emph{link} relation
is known. Unlike many other graph partitioning and community detection methods, probabilistic 
approaches can therefore easily handle incomplete graph data, where \emph{link(V,W)} atoms can 
also have an \emph{unknown} status.  

Apart from dealing with such potentially 3-valued graph data, 
we can also exploit this robustness of the likelihood function to 
improve scalability to larger networks by sub-sampling the \emph{false}-link data. 
Assuming complete network data, the number of factors in \eqref{eq:graphll}, and hence 
the number  of nodes in the likelihood graph, is quadratic in the number of nodes of the
network. Since networks tend to be sparse, the number of \emph{true} links are usually
greatly outnumbered by the \emph{false} links, and one may expect that the community structure
is already well identified by the \emph{true} links, and a random sub-sample of the 
\emph{false} links. 

To investigate the effects of learning from randomly sub-sampled data, we consider a multi-relational social 
network described in~\cite{RosMag15}. This network, which we call the \emph{Aarhus} network, contains  
61 nodes and 5 different relations. 
We apply our model \eqref{eq:singlerelmodel1},\eqref{eq:multrbnmodel} with 5 communities to data consisting of all 
the \emph{true} links (of all 5 relations), and a random sub-sample of $q\%$ ($q=100,50,20,10,5$) of the 
\emph{false} links. Having learned $\alpha_i, t_i$\ and $u$\ parameters from sub-sampled data, 
we fix the learned $t_i$\ and $u$\ parameters, and re-learn the $\alpha_i$\ parameters using the 
full data. In conjunction with these adjusted $\alpha_i$\ parameters, we evaluate the likelihood score of the
learned $t_i$\ and $u$\ parameters on the complete data. In this manner we can assess how well the 
 $t_i$,$u$\ parameters learned from sub-sampled data fit the complete data (since the learned $\alpha_i$\ 
essentially reflect estimates for the densities of the $r_i$, these cannot fit the complete data 
when learned from a sub-sample in which \emph{false} links are under-represented). 

\begin{figure}
  \centering
  \includegraphics[scale=0.4]{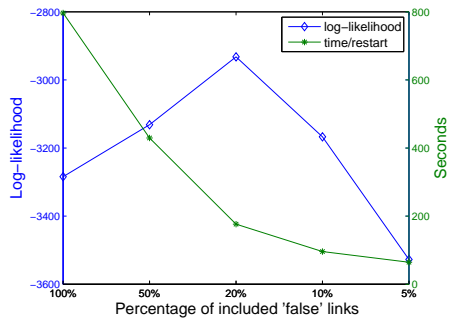}
  \caption{Learning from sub-sampled data}
  \label{fig:subsample}
\end{figure}

Figure~\ref{fig:subsample} shows the best log-likelihood score achieved in 20 restarts each for the different 
percentages of sampled \emph{false} links. Surprisingly, the likelihood score first even improves when
data is sub-sampled. A possible explanation for this can be a higher variance of the scores achieved 
in different restarts for the larger datasets, and the best out of 20 restarts being further from a 
global optimum.   Figure~\ref{fig:subsample} also shows the 
time per restart for the different data sets. These times  follow very closely the number of atoms in the 
data.  

\begin{figure}
  \centering
  \includegraphics[scale=0.4]{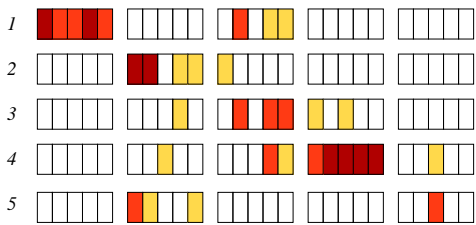}
  \caption{Pearson correlations between communities}
  \label{fig:pearson}
\end{figure}

We next try to determine how closely the communities identified from the sub-sampled data resemble the
communities found from the full data. For this we consider as a reference the communities found from
the 100\% data in an extended sequence of 38 restarts, where the best solution then obtained a 
likelihood score of -3174.  
Let $u^{\emph{ref}}(\cdot,C_k)$\ be the $u$-values for community $C_k$\ 
in this reference solution, and  $u^{q}(\cdot,C_k)$\ the $u$-values learned within 20 restarts 
from the $q\%$\ data ($q=100,50,20,10,5$). For all $q$, we compute 
Pearson's correlation between the vectors   $u^{\emph{ref}}(\cdot,C_k)$\ and $u^{q}(\cdot,C_k)$\ for 
$k=1,\ldots,5$, and then (manually) re-index the communities in $u^{q}$\ to obtain the best 
pairwise matches (according to Pearson's correlation) between the  $u^{\emph{ref}}(\cdot,C_k)$\ and $u^{q}(\cdot,C_k)$.
Figure~\ref{fig:pearson} shows a coarse heat-map visualization of all Pearson correlations
after the re-indexing. Rows in this figure correspond to the reference communities
 $u^{\emph{ref}}(\cdot,C_k)$\ ($k=1,\ldots,5$). The 5 main columns correspond to the communities
$u^{q}(\cdot,C_k)$. An element at row $k$\ and column $k'$\ consists of 5 colored rectangles, representing 
the correlation between $u^{\emph{ref}}(\cdot,C_k)$\  and $u^{q}(\cdot,C_{k'})$, for 
$q=100,50,20,10,5$\ (in this order). Dark red stands for a correlation $>0.7$, medium red for 
$>0.5$, and yellow for $>0.3$.  The result shows that reference communities 1 and 4 are usually
also identified from sub-sampled data. These two communities are also the ones which are 
identified as the most significant ones according to our significance measure, which evaluates to 
292, 183, 130, 330, 147 for communities 1,2,3,4,5, respectively.

\subsection{Related Work}

Probabilistic latent feature models for social networks (in the single-relational 
setting) have been proposed in~\cite{HofRafHan02}.
The focus there, however, is more on obtaining interpretable, visual embeddings of the
nodes in latent space, than on community analysis. 

To apply SRL modeling tools for node clustering in social network analysis has already
been suggested in~\cite{TasSegKol01}. Clusters here consist 
of nodes with similar properties, however, not of connected communities of nodes.
In~\cite{XuTreYuYu08}  a nonparametric Bayesian model with discrete
latent variables is proposed, that induces a hard partitioning of the nodes. That model is formulated
for multi-relational networks, but only applied to single-relational ones in~\cite{XuTreYuYu08}.
Similarly, an RBN model with discrete latent variables for standard partitioning-based 
community detection was presented in~\cite{JiuJae14}.

\section{Conclusion}

We have identified two distinct ways in which support for numerical data can be added to  
statistical relational models:
as numerical probabilistic relations with an associated distribution model, or as numeric input
relations, which can also be understood as object-specific model parameters. 

We have extended the RBN framework to allow for numeric input relations. Such an extension 
is particularly well supported by RBNs, because here relational (logical) atoms always have been 
treated syntactically 
and  semantically as interchangeable with numeric parameters, and only minimal adjustments to 
the language and its inference and learning algorithms are required.   
By also introducing a logistic regression combination function, we obtain a  framework
that supports standard modelling techniques for categorical data in a relational setting,
where both models and learned parameters are interpretable.  We here have focused on 
logistic regression for conditioning binary response variables on numeric inputs, but 
other models could be integrated by adding additional combination functions to the 
language. The only requirement is that the combination functions are differentiable. 

The second part of the paper applies the extended RBNs to develop new models for
community structure analysis in social networks. Specifically, we address the challenges 
of communities in multi-relational networks, and of assigning community centrality degrees
for node-community pairs. Unlike most kinds of community membership degrees that are obtained
by existing soft clustering methods, these \emph{ccd}'s are not fractional membership assignments, 
but measures for how well a node is connected with each community.
At the same time, when applied to multi-relational networks, the proposed model provides 
an explanation of how communities relate to different relations, and a validity measure that
for the significance of each detected community.

The RBN modeling tool provides a platform on which one can easily implement different 
network models for community detection, and which  are all supported by a single 
generic learning algorithm. Like for all general purpose modeling and inference tools, this 
generality comes at the price that for any particular model more efficient inference and 
learning techniques could probably be developed by dedicated implementations that can
incorporate numerous problem-specific optimizations. Thus, should at any point in time more 
``industrial strength'' applications be desired with the community structure model we 
proposed, then a new model-specific implementation may be needed. 

With the network models we have investigated in this paper we have stayed close to 
established models, and only scratched the surface of the modeling capabilities provided
within the RBN language. One line of future work is to integrate these structural network models with
dynamic models for information diffusion within the networks. 

The computational bottleneck in our implementation at this point is the size of the 
likelihood graph. We have already shown that to some extent this problem can be reduced 
by sub-sampling the false edges of the network. Other techniques we are currently exploring
are optimization strategies in which in an iterative manner the likelihood function is only 
partially optimized based on smaller,  partial likelihood graphs.  Such iterative 
partial optimizations can either follow a block gradient descent strategy, in which only
subsets of parameters are optimized in each iteration, stochastic gradient descent strategies,
in which only the likelihood function of a part of the data is optimized, or a combination
of both these approaches. The challenge is to develop generic strategies that are widely
applicable to a broad range of models, and that do not require the user to perform 
model-specific tuning of the learning strategy in each case.

In principle, it would also be quite straightforward to add models for numeric
probabilistic relations to the RBN framework. The language of probability formulas 
can directly be used also to define mean and variance of a Gaussian distribution (for example), 
and thereby  define Gaussians that are in complex ways conditioned on 
continuous and categorical predictors. 
However, this will come at the price of
loosing the tools for exact inference, and one would have to rely on sampling-based
inference methods. 

All results  presented in this paper are obtained
using an updated version of the \emph{Primula} implementation of 
RBNs\footnote{\url{ http://people.cs.aau.dk/~jaeger/Primula}}, which will become
available with then next system release.

\bibliographystyle{plain}

\begin{thebibliography}{10}

\bibitem{Agresti02}
A.~Agresti.
\newblock {\em Categorical Data Analysis}.
\newblock Wiley, 2002.

\bibitem{BerPinCal13}
Michele Berlingerio, Fabio Pinelli, and Francesco Calabrese.
\newblock Abacus: frequent pattern mining-based community discovery in
  multidimensional networks.
\newblock {\em Data Mining and Knowledge Discovery}, 27(3):294--320, 2013.

\bibitem{BreBooAra75}
Ronald~L Breiger, Scott~A Boorman, and Phipps Arabie.
\newblock An algorithm for clustering relational data with applications to
  social network analysis and comparison with multidimensional scaling.
\newblock {\em Journal of Mathematical Psychology}, 12(3):328--383, 1975.

\bibitem{CaietAl05}
Deng Cai, Zheng Shao, Xiaofei He, Xifeng Yan, and Jiawei Han.
\newblock Mining hidden community in heterogeneous social networks.
\newblock In {\em Proceedings of the 3rd international workshop on Link
  discovery}, pages 58--65. ACM, 2005.

\bibitem{DeRFraKerMug08}
L.~De~Raedt, P.~Frasconi, K.~Kersting, and S.H. Muggleton, editors.
\newblock {\em Probabilistic Inductive Logic Programming}, volume 4911 of {\em
  Lecture Notes in Artificial Intelligence}.
\newblock Springer, 2008.

\bibitem{DomRic01}
Pedro Domingos and Matt Richardson.
\newblock Mining the network value of customers.
\newblock In {\em Proceedings of the seventh ACM SIGKDD international
  conference on Knowledge discovery and data mining}, pages 57--66. ACM, 2001.

\bibitem{GetTas07}
L.~Getoor and B.~Taskar, editors.
\newblock {\em Introduction to Statistical Relational Learning}.
\newblock MIT Press, 2007.

\bibitem{GirNew02}
Michelle Girvan and Mark~EJ Newman.
\newblock Community structure in social and biological networks.
\newblock {\em Proceedings of the National Academy of Sciences},
  99(12):7821--7826, 2002.

\bibitem{GutJaeDeR10}
B.~Gutmann, M.~Jaeger, and L.~De Raedt.
\newblock Extending problog with continuous distributions.
\newblock In {\em Proceedings of the 20th Int. Conf. on Inductive Logic
  Programming (ILP)}, volume 6489 of {\em LNCS}, pages 76--91. Springer, 2011.

\bibitem{HofRafHan02}
Peter~D Hoff, Adrian~E Raftery, and Mark~S Handcock.
\newblock Latent space approaches to social network analysis.
\newblock {\em Journal of the American Statistical Association},
  97(460):1090--1098, 2002.

\bibitem{Hofmann04}
Thomas Hofmann.
\newblock Latent semantic models for collaborative filtering.
\newblock {\em ACM Transactions on Information Systems (TOIS)}, 22(1):89--115,
  2004.

\bibitem{Jaeger97UAI}
M.\marginlabel{*} Jaeger.
\newblock Relational bayesian networks.
\newblock In Dan Geiger and Prakash~Pundalik Shenoy, editors, {\em Proceedings
  of the 13th Conference of Uncertainty in Artificial Intelligence (UAI-13)},
  pages 266--273, Providence, USA, 1997. Morgan Kaufmann.

\bibitem{JaegerICML07}
M.~Jaeger\marginlabel{*}.
\newblock Parameter learning for relational {B}ayesian networks.
\newblock In {\em Proceedings of the 24th International Conference on Machine
  Learning (ICML)}, 2007.

\bibitem{JiuJae14}
J.~Jiang and M.~Jaeger.
\newblock Community detection for multiplex social networks based on relational
  bayesian networks.
\newblock In {\em Proc. of the 21st Int. Symposium on Methodologies for
  Intelligent Systems (ISMIS)}, 2014.

\bibitem{KivetAl14}
Mikko Kivel{\"a}, Alex Arenas, Marc Barthelemy, James~P Gleeson, Yamir Moreno,
  and Mason~A Porter.
\newblock Multilayer networks.
\newblock {\em Journal of Complex Networks}, 2(3):203--271, 2014.

\bibitem{Lee1974}
Elisa~T Lee.
\newblock A computer program for linear logistic regression analysis.
\newblock {\em Computer programs in biomedicine}, 4(2):80--92, 1974.

\bibitem{RavRamDav15}
I.~Ravkic, J.~Ramon, and J.~Davis.
\newblock Learning relational dependency networks in hybrid domains.
\newblock {\em Machine Learning}, 2015.
\newblock to appear.

\bibitem{RicDom06}
M.~Richardson and P.~Domingos.
\newblock {M}arkov logic networks.
\newblock {\em Machine Learning}, 62(1-2):107 -- 136, 2006.

\bibitem{RosMag15}
Luca Rossi and Matteo Magnani.
\newblock Towards effective visual analytics on multiplex and multilayer
  networks.
\newblock {\em Chaos, Solitons \& Fractals}, 72:68--76, 2015.

\bibitem{TasSegKol01}
Benjamin Taskar, Eran Segal, and Daphne Koller.
\newblock Probabilistic classification and clustering in relational data.
\newblock In {\em Proc. of 17th International Joint Conference on Artificial
  Intelligence}, pages 870--878, 2001.

\bibitem{WanDom08}
J.~Wang and P.~Domingos.
\newblock Hybrid markov logic networks.
\newblock In {\em Proceedings of the Twenty-Third National Conference on
  Artificial Intelligence}, 2008.

\bibitem{XuYurFenSch07}
Xiaowei Xu, Nurcan Yuruk, Zhidan Feng, and Thomas~AJ Schweiger.
\newblock Scan: a structural clustering algorithm for networks.
\newblock In {\em Proceedings of the 13th ACM SIGKDD international conference
  on Knowledge discovery and data mining}, pages 824--833. ACM, 2007.

\bibitem{XuTreYuYu08}
Zhao Xu, Volker Tresp, Shipeng Yu, and Kai Yu.
\newblock Nonparametric relational learning for social network analysis.
\newblock In {\em KDD~2008 Workshop on Social Network Mining and Analysis},
  2008.

\bibitem{Zachary77}
W~Zachary.
\newblock An information flow modelfor conflict and fission in small groups1.
\newblock {\em Journal of anthropological research}, 33(4):452--473, 1977.

\end{thebibliography}

\end{document}